\documentclass[runningheads]{llncs}

\usepackage[mobile]{eccv}

\usepackage{eccvabbrv}

\usepackage{graphicx}
\usepackage{booktabs}

\usepackage[accsupp]{axessibility}  

\usepackage[utf8]{inputenc} 
\usepackage[T1]{fontenc}    
\usepackage{url}            
\usepackage{booktabs}       
\usepackage{amsfonts}       
\usepackage{nicefrac}       
\usepackage{microtype}      
\usepackage{xcolor}         
\usepackage{wrapfig}
\usepackage{graphicx}    
\usepackage{multirow}
\usepackage{tabularx}
\usepackage{colortbl}
\usepackage{amsmath}
\definecolor{Gray}{gray}{0.9}
\definecolor{Gray2}{gray}{0.95}
\usepackage{pifont}
 \newcommand{\cmark}{\ding{51}}%
\newcommand{\xmark}{\ding{55}}%

    \PassOptionsToPackage{numbers, compress}{natbib}

\newcommand{\tablestyle}[2]{\setlength{\tabcolsep}{#1}\renewcommand{\arraystretch}{#2}\centering\small}
\newlength\savewidth

\usepackage[breaklinks,colorlinks,citecolor=eccvblue]{hyperref}

\usepackage{orcidlink}

 		\footnotetext[1]{Most work done during internship at King Abdullah University of Science and Technology. }

\begin{document}

\title{How Well Can Vision Language Models See \\
Image Details?} 

\author{Chenhui Gou \inst{1}\textsuperscript{*} \and
Abdulwahab Felemban \inst{2} \and
Faizan Farooq Khan \inst{2} \and \\
Deyao Zhu \and \inst{2} Jianfei Cai \inst{1} \and Hamid Rezatofighi \inst{1}  \and Mohamed Elhoseiny \inst{2}}


\institute{Monash University \and
King Abdullah University of Science and Technology}

\maketitle

\begin{abstract}
Large Language Model-based Vision-Language Models (LLM-based VLMs) have demonstrated impressive results in various vision-language understanding tasks. However, how well these VLMs can see image detail beyond the semantic level remains unclear. In our study, we introduce a pixel value prediction task (PVP) to explore "How Well Can Vision Language Models See Image Details?" and to assist VLMs in perceiving more details.
Typically, these models comprise a frozen CLIP visual encoder, a large language model, and a connecting module. After fine-tuning VLMs on the PVP task, we find: 1) existing VLMs struggle to predict precise pixel values by only fine-tuning the connection module and LLM; and 2) prediction precision is significantly improved when the vision encoder is also adapted. Additionally, our research reveals that incorporating pixel value prediction as one of the VLM pre-training tasks and vision encoder adaptation markedly boosts VLM performance on downstream image-language understanding tasks requiring detailed image perception, such as referring image segmentation (with an average +10.19 cIoU improvement) and video game decision making (with average score improvements of +80.34 and +70.54 on two games, respectively).

\end{abstract}

\section{Introduction}
\label{sec:intro}

Large Language Models (LLMs) have revolutionized the field of artificial intelligence, enabling machines to perceive and generate human-like text with remarkable performance. Following this advancement, LLM-based Vision-Language Models (VLMs) are rapidly evolving within the cross-domain of vision and language. Recent VLMs, such as ~\cite{minigpt4,minigptv2,llava,llava1.5} have shown promising performance on multiple vision-language tasks, including visual question answering (VQA) and referring expression comprehension (REC). Typically, these LLM-based VLMs adopt a similar modeling design: a pre-trained visual encoder to extract visual features, a projection module to align these features to the language space, and an LLM to perform reasoning. 
\begin{figure}[!t]
    \centering
    \includegraphics[scale=0.55]{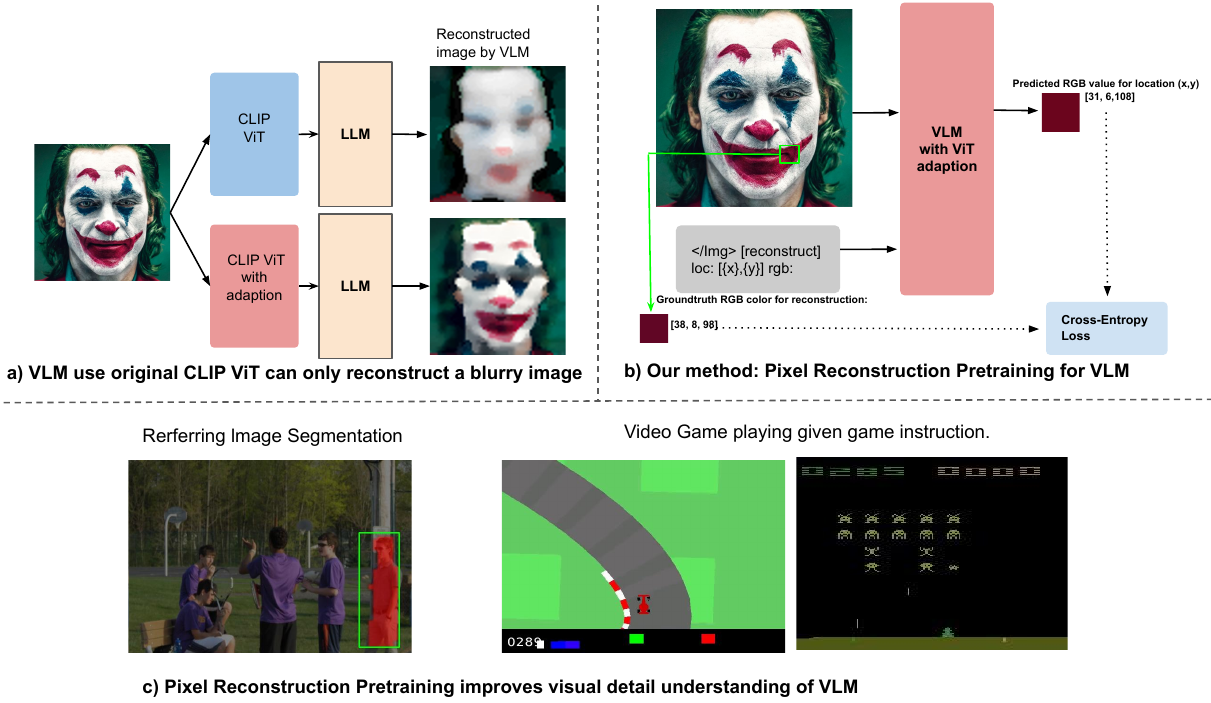}
    \caption{\textbf{Method.} a) shows our findings: 
Using the original CLIP vision features, VLMs can only reconstruct a blurry contour without many visual details. The reconstruction result can be improved by adapting the vision encoder. The reconstructed image is generated by querying pixel values with pixel locations, as shown in (b). For better illustration, the connection module between ViT and LLM is ignored. b) shows that we incorporate pixel prediction as a pretraining task for VLM. c) illustrates some downstream tasks performed by VLM, which require both vision detail understanding and language information. Our pretraining improves VLM performance on these tasks.}
    \label{fig:tile_image}
    
\end{figure}

These VLMs primarily utilize CLIP-based (Contrastive Language-Image Pre-training) visual encoders (e.g., CLIP~\cite{radford2021learning}, OpenCLIP~\cite{openclip}, EVA-CLIP~\cite{fang2023eva} and CLIP image features are advantageous for interpretation by LLMs ~\cite{merullo2022linearly} because these features are aligned with the language space through training on large-scale image-text paired datasets.
However, as CLIP image features are aligned with short and brief language captions, it remains uncertain whether these LLMs can truly "see" the original image content. 

To investigate this, we propose a method to show how well current VLMs can perceive visual details from original CLIP vision features by examining their ability to predict pixel values of the perceived images. Inspired by Implicit Neural Representation (INR) in image generation~\cite{tancik2020fourier, sitzmann2020implicit, anokhin2021image, skorokhodov2021adversarial, haydarov2022hypercgan}, we design a pixel value prediction (PVP) task in a visual question-answering format, which can be directly integrated into existing VLM pipelines. Given image CLIP features and an (x, y) coordinate, we prompt the Large Language Model (LLM) to predict the RGB pixel value at that coordinate. The question format can be seen in~\cref{fig:tile_image}b). We first fine-tune the VLM following common protocol, training the connection module and LLM while freezing the vision encoder.
As shown in~\cref{fig:tile_image}a), for better visualization, we visualize the reconstructed image by querying all pixel locations in batch inference, while during training we only randomly sample a location from a random image in the training set. We find that VLMs with a frozen CLIP encoder can only reconstruct a blurry contour without many visual details. Furthermore, we notice a significant improvement in pixel prediction results if we also adapt the CLIP encoder for the PVP task. More visualization results can be viewed in~\cref{fig:reconstruction}. 
Also, Pixel reconstruction~\cite{he2022masked, gao2022convmae, hu2022exploring} is a classical and effective vision pre-training task for downstream tasks that require an understanding of vision details, such as segmentation or depth estimation~\cite{kirillov2023segment, he2022masked, jepa}. Inspired by our findings and the previous successes of the pixel reconstruction task transferred to detailed vision tasks, we adapt pixel value prediction as a pre-training task for Vision-Language Models (VLMs), as demonstrated in~\cref{fig:tile_image}b. and expect the enhanced perception ability to be helpful for downstream tasks that require detailed vision and language understanding. Due to the special properties of our pre-trained model, we refer to it as the Pixel Autoencoded Large MultiModal Model (PAE-LaMM).

To validate whether the improved pixel prediction ability can truly help better vision detail understanding ability in VLMs, we selected two downstream vision-language tasks that require visual details to compare the performance of the base VLM and PAE-LaMM: the Referring Image Segmentation task~\cite{kazemzadeh2014referitgame, nagaraja2016modeling} and the video game playing task, as shown in~\cref{fig:tile_image}c. In the segmentation task, VLMs need to accurately perceive the shape of an object referenced in a given phrase within an image and generate its segmentation mask. For video game playing tasks such as Car Racing and Space Invaders, VLMs need to correctly interpret visual elements like the road or enemy bullets, generating appropriate actions based on stacked video frames and the game description. We collected datasets comprising over 53K observation-action pairs played by expert reinforcement learning models and trained our model to imitate the actions of these experts.
In the experiments section, we first present the performance gap on the PVP task between fine-tuning VLM with a frozen vision encoder and ViT adaptation. Next, we illustrate how our pre-training task and vision encoder adaptation strategy benefit downstream tasks like referring segmentation and video gaming through improved perception of visual details. Finally, we show our method's performance on mainstream Visual Question Answering (VQA) tasks, demonstrating that our pre-trained model achieves results comparable to state-of-the-art approaches while also offering additional capabilities in pixel reconstruction.
Our contribution can be summarized as follows:
\begin{itemize}
\setlength{\itemsep}{2pt}
\item We propose a pixel value prediction (PVP) task to examine the ability of current LLM-based Vision-Language Models in perceiving original image details. This task is designed as a vision question-answer type and can be easily integrated into existing VLM pipelines without additional design.
\item By fine-tuning VLMs on the PVP task, our research shows that these models face challenges in accurately discerning pixel-level details. Performance significantly improves when adapting their typically frozen vision encoder, revealing that the frozen CLIP vision encoder limits these VLMs in perceiving visual details.
\item We incorporate PVP into the existing VLM pre-training pipeline and adapt the vision encoder during training. Results show our pretraining helps VLMs perform better in downstream vision-language tasks requiring the perception of visual details, such as image segmentation and video game decision making.
\end{itemize}


\section{Related Work}
\textbf{LLM-based Vision-Language Models.} 
Recent advancements in Large Language Model-based Vision-Language Models (LLM-VL models) have demonstrated remarkable achievements in tasks requiring both visual comprehension and language understanding~\cite{minigpt4, minigptv2, llava1.5}. The effectiveness of LLM-based VLMs largely stems from the reasoning and generalization capabilities of Large Language Models~\cite{chatgpt, llama, llama2, vicuna2023} trained on large-scale datasets. Recent studies have explored the abilities of Large VLMs in visual grounding tasks~\cite{shikra, wang2024visionllm, minigptv2} and referring image segmentation~\cite{lai2023lisa, ren2023pixellm}. However, these works mainly focus on aligning different levels of semantic vision information and language. How these Vision-Language Models interpret the original image and whether they can see the original image details beyond semantic information is less investigated. To address this gap, we propose a method to investigate the original vision detail perception ability and design a self-supervised pre-training method to augment their original image perceptual capabilities. We validate that the enhanced ability can boost performance in many downstream tasks requiring detailed vision and language understanding.

\noindent\textbf{Pixel Reconstruction as Pretraining.}
Image pixel reconstruction has been explored as an effective method for pre-training computer vision models~\cite{he2022masked,gao2022convmae,hu2022exploring}. Pre-training vision models with a reconstruction task also aid in vision-specific tasks that require pixel-level understanding such as semantic segmentation~\cite{he2022masked}, class-agnostic segmentation~\cite{kirillov2023segment} and depth estimation~\cite{jepa}. 
However, current LLM-based VLM models primarily utilize vision encoders pre-trained through vision-language contrastive learning, as these features can be well understood by the Large Language Model ~\cite{merullo2022linearly}. Thus, it is less effective to simply plug a vision encoder into VLM that is pre-trained separately on a reconstruction task. Additionally, it remains unclear how to incorporate the reconstruction task into the training of VLM and whether it will enhance the entire VLM's understanding of visual details. Considering the general vision-language task paradigm, we design the pixel reconstruction as the VQA task and update the entire VLM on this task to improve visual detail understanding, rather than focusing solely on the vision model.
\\
\textbf{VLMs on Referring Image segmentation}.
Referring Image segmentation task~\cite{kazemzadeh2014referitgame,nagaraja2016modeling} aims to segment a specific object based on a given sentence description. This task requires pixel-level vision detail and language understanding. VisionLLM \cite{wang2024visionllm} considers segmentation masks as polygons sequence prediction while needing expanded vocabulary for LLM decoder and extra vision decoder for image tokenizer. Lisa~\cite{lai2023lisa} combines LLM and strong segmentation expert SAM ~\cite{kirillov2023segment} to do complex instruction reasoning segmentation. In our method, we do not involve any extra vision component besides the CLIP encoder, and referring segmentation results are predicted directly from LLM. We show the VLM can provide a precise pixel-level mask if it can see image detail better.\\
\textbf{Video Games Playing by Large Language Models}.
Games play a crucial role in AI research, requiring multiple abilities from AI models, such as high-level planning and reasoning~\cite{yannakakis2018artificial}. Recently, LLMs have been investigated for their potential as player agents in many game applications due to their excellent reasoning ability~\cite{wang2023voyager, tsai2023can, akata2023playing, wu2024read}. Similarly, humans can play video games by watching videos and understanding the game's instructions. A recent work ~\cite{wu2024read} utilized human-written instructions and LLMs to accelerate Reinforcement Learning (RL) algorithms for Atari games. Video games, in particular, require both vision perception from videos and language understanding of game instructions. Specifically, some video games necessitate detailed image analysis, such as Carracing~\cite{klimov2016carracing} and the Atari game Space Invaders. For instance, a model may need to control steering as the car approaches a corner on the driving road.
Thus, we use VLM to play video games and show that VLM can achieve a higher score if the vision perception ability is improved.
\section{Method}

We first introduce our method for investigating the image perception abilities of current Vision-Language Models (VLMs). We then present the design of our pre-training task, pixel reconstruction for Large Language Model (LLM)-based VLMs. Following this, we outline the designs of our downstream tasks, including referring image segmentation and video game playing.

\subsection{Method for investigating image perception ability of VLMs.}

We begin by examining the ability of Vision-Language Models (VLMs) to understand image details by engaging them in pixel reconstruction tasks, which require the model to perceive images at the pixel level. To adapt this task for VLMs, we conceptualize pixel reconstruction as a Visual Question Answering (VQA) task. We prompt the VLM to give the pixel value at a specific location (x,y) on the image, as illustrated in part b) of~\cref{fig:tile_image}. In line with the task design for Large Language Model (LLM)-based VLMs, we introduce a task identifier \textit{[reconstruct]} followed by the question:
\begin{center}
\textit{$<$Img$>$ $<$ ImageFeature$>$ $<$/Img$>$ [reconstruct] loc: [\{x\},\{y\}] rgb: \}}
\end{center}
The answer format is \textit{[${r},{g},{b}$]}, where $r$, $g$, and $b$ represent the RGB values, respectively. 
Fine-tuning the current VLM for this task reveals that the LLM can reconstruct only a blurry image when using vision embedding from original CLIP, while only training the Large Language Model and the connection module according to previous training paradigms.
We find that the quality of pixel reconstruction greatly improves when we also adapt the vision encoder during the training process. The comparison results are displayed in~\cref{fig:reconstruction}.

\subsection{Pixel Reconstruction Pre-training for VLMs}

We incorporate PVP into Vision-Language Model (VLM) pretraining pipeline. Our method aims to enhance the ability of current VLMs to understand detailed visual information without losing general vision-language knowledge. We introduce pixel reconstruction as a new task for Visual Question Answering (VQA) and include it, along with other vision-language tasks, in our model's training. For the additional tasks, we follow the complete set from the instructional training of MiniGPTv2~\cite{minigptv2}, and we use the same datasets as those in~\cite{minigptv2}. We have created a three-stage training approach to help our model better align with visual details.

The first stage aims to familiarize the VLMs with new pixel reconstruction tasks. Following the previous training settings used in~\cite{minigptv2}, we train only the Large Language Model (LLM) and the connection modules in this stage, as we find that directly unfreezing the vision encoder leads to catastrophic forgetting (experiments are provided in the supplementary materials). In the second stage, in addition to the LLM and connection modules, we adapt the vision encoder to improve the VLM's ability to understand visual details. In the last stage, we freeze the vision encoder and reduce the sampling ratio for the pixel reconstruction task to achieve a balance between lower-level details and high-level semantics in the vision-language space. We utilize LoRA to efficiently train the LLM. After three stages of pretraining, we evaluate our approach on downstream tasks such as Referring Image Segmentation and Video Game Playing. Necessary details of the pretraining settings are provided in the supplementary materials.
\subsection{Referring Image Segmentation}
Following a similar paradigm as pixel reconstruction, we consider Referring to Image Segmentation as a VQA task. We ask VLM to provide the answer to the question "Does this pixel location $(x,y)$ contain a specific object described in the referring sentence?" We use a task identifier [segmentation] and follow the question:

\begin{center}
\textit{$<$Img$>$ $<$ ImageFeature$>$ $<$/Img$>$ [segmentation] \{referring sentence\} loc: [\{x\},\{y\}] mask: \}}
\end{center}

The answer would be \textit{${0}$} or \textit{${1}$}, and $0,1$ represent the binary mask of this object at location $(x,y)$. We don't use an extra decoder or special codebook for segmentation. The prediction mask is directly generated by the Large Language Model in VLM by examining vision features and language guidance. The performance of Referring Image Segmentation can reflect the pixel-level vision understanding ability of the VLM.
\begin{figure}[!t]
    \centering
    \includegraphics[scale=0.45]{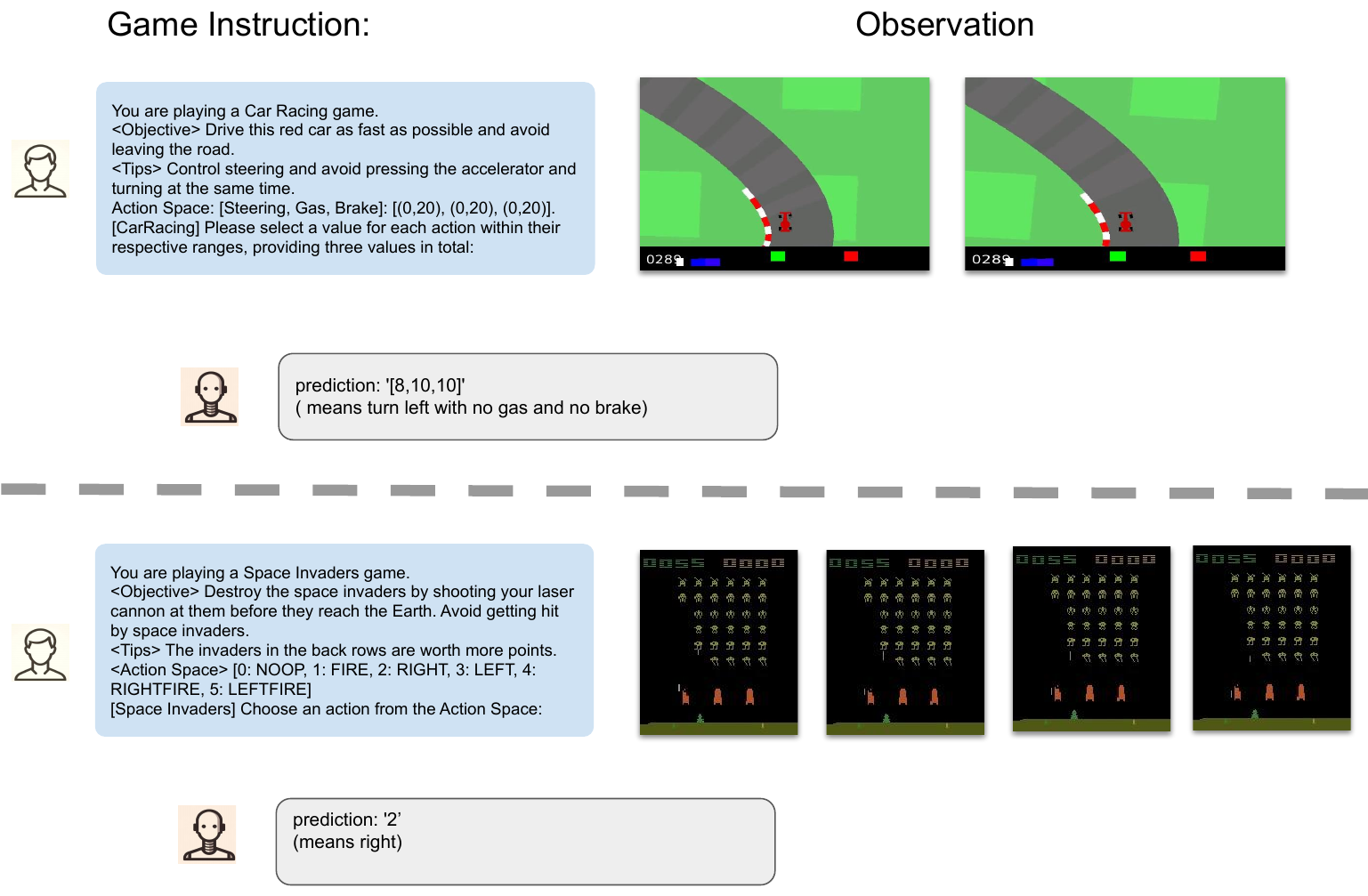}

    \caption{\textbf{Examples of Game Playing by VLM.} The input to the VLM is the stacked images and the game instructions. The first row shows an example of playing Carracing. The second row shows the SpaceInvaders game. The number of stacked frames depends on the expert model we used. For example, Carracing uses two frames and SpaceInvaders uses four.}
    \label{fig:illustration:gaming}
    
\end{figure}
\subsection{Video Games Playing}
We pick two video games sourced from OpenAI Gym environment~\cite{brockman2016openai}: Carracing Game and SpaceInvaders. We consider playing a Video Game as a Video Question Answering task, where each short video contains \textit{N} stacked images. Given one observation (one short video), the model needs to predict actions based on the action space of each game. We first design a general template for video game playing in the following format:
\begin{center}
\textit{\small$<$Img$>$ $<$Image1Feature$>$ $<$/Img$>$$<$Img$>$ $<$ Image2Feature$>$ $<$/Img$>$ ...  $<$ImageNFeature$>$ $<$/Img$>$ \{game instruction\} [game identify] choose an action from Action Space: \}}
\end{center}
\textit{ImageNFeature} is the vision features of $N_{th}$ image. 
Game instruction is the necessary information for playing this game, which contains the objective of the player, game tips, and the action space of the game.~\cref{fig:illustration:gaming} shows the illustration of gaming playing using LLM-based VLM. We find one pre-trained expert Reinforcement Learning (RL) model for each game provided by stable-baselines3~\cite{stable-baselines3} according to game scores. We collect the dataset consisting of observations and corresponding actions taken by the expert model. During testing, we set the game seed to be different from the training environment.
We consider the fine-tuning process of VLM on game playing as imitation learning, while we only use the same loss used in other VQA tasks. The output is directly generated by the VLM without any interpretation or extra decoders. \\
\textbf{Carracing Game.}
We choose CarRacing-v0, as it is a widely used version. We choose RecurrentPPO as the expert model provided by stable-baselines3~\cite{stable-baselines3}.
This model uses two stacked frames as single observation input, and takes an action which is a vector containing three continuous values ($C_1,C_2,C_3$) representing (steering, gas, brake) respectively. We first map these three values to the discrete values, detail is provided in supplementary material. The data we collected is the observation and the action value taken by this expert model of each step. We collect a dataset containing $30$ games (with different game seeds), in total $28585$ observations (stacked images), and corresponding actions.
\\\textbf{SpaceInvaders.}
We use the SpaceInvadersV4 version and choose a pre-trained DQN agent to play this game. This model uses four stacked frames as a single observation input and chooses one action from the following action space ([0: NOOP, 1: FIRE, 2: RIGHT, 3: LEFT, 4: RIGHTFIRE, 5: LEFTFIRE]). We directly document each observation and the action taken by the expert model. $30$ games are collected, containing $24618$ observations and corresponding actions.

\noindent\section{Experiments}
In this section, we present experimental settings and results. We first show the reconstruction results of baseline VLM and our method. Then we report our results on two types of downstream tasks to demonstrate how much benefit VLM can get from pixel reconstruction pretraining.
The first one is referring image segmentation and the second one is video game playing. We demonstrate both quantitative and qualitative results.
In the last section, we show our pre-trained model can also achieve comparable results on other vision Language tasks and owns an extra pixel reconstruction ability. The ablation study for the pre-training strategy is provided.
\begin{table*}[t!]
\caption{The training datasets used for our Pre-training.
}

\centering 
\resizebox{\textwidth}{!}{
\begin{tabular}{@{}ll@{}} %
\toprule
 Data types &   Dataset \\ 
 \midrule
Reconstruction & COCO caption ~\cite{lin2014microsoft} \\
Caption &  COCO caption ~\cite{lin2014microsoft},  Text Captions ~\cite{sidorov2019textcaps} \\
REC &  RefCOCO~\cite{kazemzadeh2014referitgame}, RefCOCO+~\cite{yu2016modeling}, RefCOCOg~\cite{mao2016generation}, Visual Genome ~\cite{krishna2017visual}  \\
REG & RefCOCO~\cite{kazemzadeh2014referitgame}, RefCOCO+~\cite{yu2016modeling}, RefCOCOg~\cite{mao2016generation}  \\
VQA &  GQA~\cite{hudson2019gqa}, VQAv2~\cite{goyal2017making}, OCR-VQA~\cite{ocrvqa}, OK-VQA~\cite{marino2019ok}, AOK-VQA~\cite{schwenk2022okvqa} \\
Multimodal instruction & LLaVA dataset~\cite{llava}, Flickr30k~\cite{flickr30k}, Multi-task conversation~\cite{minigptv2}\\
Language dataset & Unnatural Instructions~\cite{honovich2022unnatural}\\

\bottomrule
\end{tabular}
}

\label{tab:three_stages_dataset}
\end{table*}\\
\noindent\textbf{Implementation details.}  
We use MiniGPT-v2~\cite{minigptv2}  as our VLM base and utilize their pre-trained weights to initialize our model. To investigate how adapting the vision encoder affects the performance of VLM on the pixel prediction task, we first obtain a VLM trained with the first stage of pretraining introduced in our method. Then, we continue to train the model using two strategies: freezing the ViT and adapting the vision encoder. We compare the qualitative and quantitative image reconstruction performance of these two models after the second stage to measure the pixel prediction quality. For pretraining details, we employ the three-stage training strategy. Following a similar setting as described in~\cite{minigptv2}, we utilize LoRA~\cite{lora} to accelerate our training. The entire pretraining stage involves training the Large Language Model and the connection module via low-rank adaptation, with the LoRA rank set to 64. The vision encoder is only adaptable in the second stage, without using LoRA. The input image resolution is $448\times448$. The reconstruction target is the downsampled image at a resolution of $64\times64$. The complete dataset used in pretraining is shown in~\cref{tab:three_stages_dataset}. The entire pretraining stage comprises approximately 3.6M pixel reconstruction questions by randomly sampling locations from images in the COCO caption dataset~\cite{lin2014microsoft}. For downstream tasks, we directly fine-tune our baseline model and our model after three-stage pretraining, the PAE-LaMM model, and report their performance. In the referring segmentation task, we fine-tune both models on Referring Expression Comprehension (REC) and referring segmentation data. Consequently, our fine-tuned model acquires both localization and pixel-level understanding capabilities. As for video game playing, we utilize the official game environment sourced from the OpenAI Gym library~\cite{openaiGym} and employ RL-Zoo3~\cite{rl-zoo3} for data collection and as the game interface for inference. \\
\textbf{Training and Hyperparameters.} We utilize a cosine learning rate and the AdamW optimizer to train our model. All models are trained on 4xA100 GPUs. The batch size of the pixel reconstruction task is 64, 16, and 64 in each stage, respectively.
For Referring Image Segmentation, we set the batch size to 64 for segmentation and 24 for localization. For CarRacing and Space Invaders, we set the batch sizes to 8 and 3, respectively. More detailed hyperparameters will be provided in the supplementary material.
\begin{figure}[!t]
    \centering
    \includegraphics[scale=0.4]{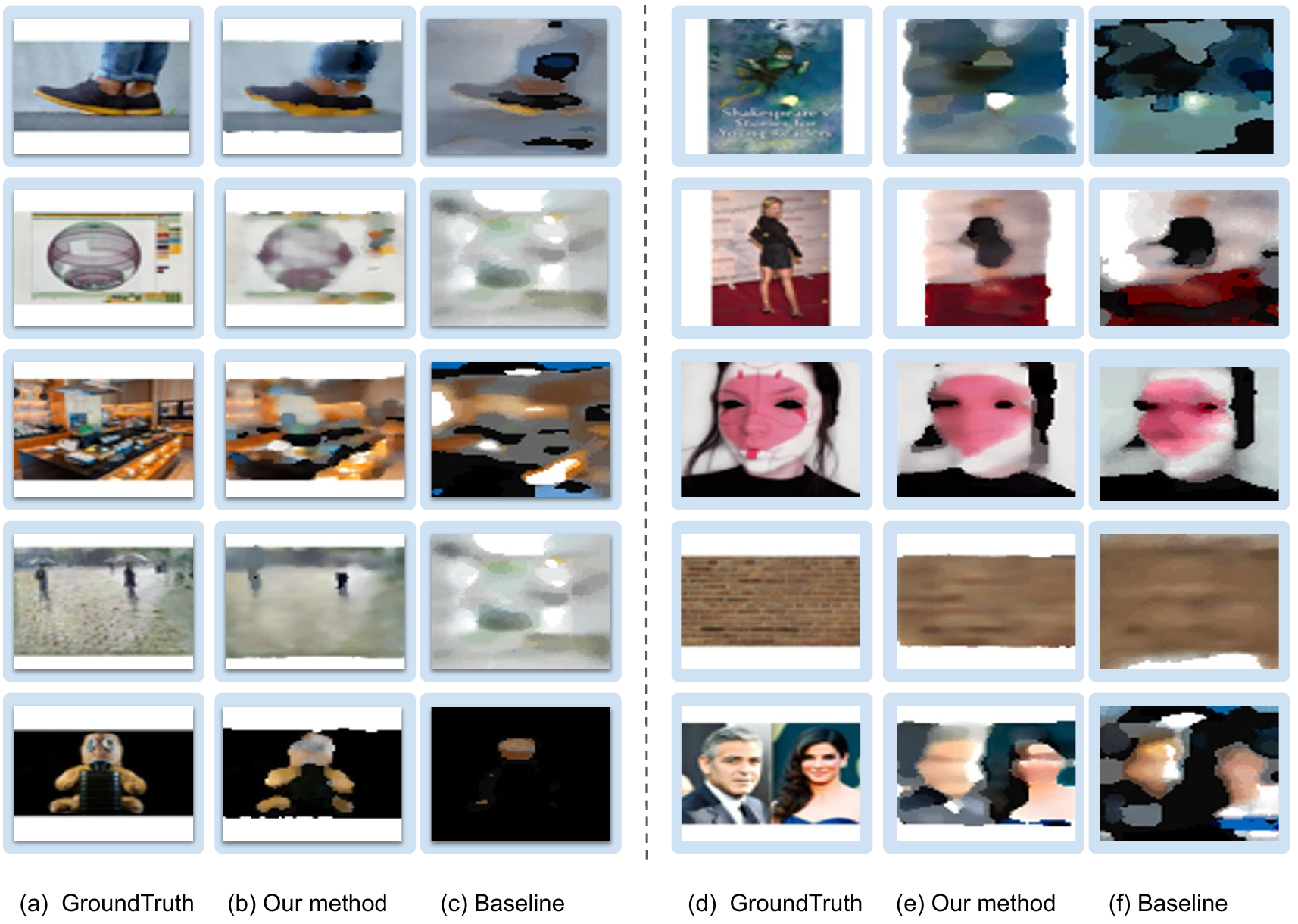}

   \caption{\textbf{Qualitative results of Reconstruction} (a) and (d) are the  GroundTruth for reconstruction. 
  (b) and (e) is the reconstructed image of our method. (c) and (f) are the baseline result without CLIP-Vit adaptation. Compared with the baseline, our method reconstructs images with more details. The averaged Reconstruction error of our method and baseline on these 10 images are $6.67$, and $24.56$, respectively.}
    \label{fig:reconstruction}
    \vspace{-3mm}
\end{figure}
\setlength{\tabcolsep}{4pt}
\begin{table}[t]
\caption{\textbf{Abliation study for adapting Vit.} We report the Reconstruction Error, average referring expression comprehension (REC)
on refcoco, refcoco+ and refcocog. Also, we report the VQA task performance. The best performance for each benchmark is indicated in \textbf{bold}. We report top-1 accuracy for other VQA tasks: GQA~\cite{hudson2019gqa}, VSR~\cite{liu2023visual}, IconVQA~\cite{iconqa} and VizWiz~\cite{vizwiz}}\label{tab:abilation:adaptVit}
\centering 
\resizebox{ \textwidth}{!}{
\begin{tabular}{@{}c c c|c cccc @{}} %
\toprule

adapt ViT & RE $\downarrow$ & Average REC $\uparrow$& GQA$\uparrow$ & VSR $\uparrow$ & IconVQA $\uparrow$& VizWiz $\uparrow$& HM $\uparrow$ \\  

\midrule

\xmark   & 20.38 & 77.2 &55.5 & \textbf{56.7} &\textbf{49.7}  &\textbf{ 53.7} & \textbf{57.6} \\
\cmark &\textbf{6.65} &\textbf{82.3} & \textbf{56.2} & \textbf{56.7} & 49.6 & 53.22 & \textbf{57.6} \\

\bottomrule
\end{tabular}
}

\end{table}
\setlength{\tabcolsep}{1.4pt}

\subsection{Evaluation on pixel reconstruction}
~\cref{fig:reconstruction} 
shows the qualitative comparison results between using our method and the baseline. The baseline model can only reconstruct a blurry contour without many visual details, indicating that the VLM cannot see enough original image detail from the original CLIP vision feature. In contrast, our method helps the VLM reconstruct a better result with more detail. Additionally, we use mean reconstruction error (RE) to report the quantitative result of pixel reconstruction as shown in the following formula: 
\begin{align}
\text{RE}(x,y) = \left| \text{pt\_rgb}(x,y) - \text{gt\_rgb}(x,y)\right| ; \\
\text{RE}(I) = \left(\frac{\sum_{x=0}^{W}\sum_{y=0}^{H} \text{RE}(x,y)}{H \times W \times 255}\right)
\end{align}
where $\text{RE}(x,y)$ is the reconstruction error for a single pixel, and $\text{RE}(I)$ is the summed error across all locations of the entire image $I$. Here, $pt\_rgb(x,y)$ represents the reconstruction prediction at location $(x,y)$, and $gt\_rgb(x,y)$ is the ground truth pixel value.  $H,W$ are the height and width of reconstruction image, and $255$ is the normalization factor.
The evaluation set comprises $409,600$ pixel reconstruction questions sampled from images in the test set of Conceptual Captions~\cite{sharma2018conceptual} and requires the VLM to provide the RGB value for each question.~\cref{tab:abilation:adaptVit} shows the quantitative results on this evaluation set; our method achieves a lower average RE and significantly outperforms the VLM without adaptation, $6.65$ vs $20.38$. This demonstrates that our method enhances the VLM's perception of visual details, enabling it to reconstruct better images. Additionally, we observed that during adaptation, the model does not lose its general vision-language knowledge, as reflected in other VQA tasks. Furthermore, the performance in referring expression comprehension is improved, attributed again to the enhanced awareness of visual details.
\vspace{-4mm}
\subsection{Results on Downstream tasks}
After pretraining, we validate the effectiveness of our method on downstream tasks that require visual detail and language understanding.
\begin{table*}[t!]
\scriptsize
\centering

\caption{
    \label{tab:only2seg}
    \textbf{Comparison result on referring segmentation.} For each dataset, we test 100 referring sentences.
    We report the official evaluation metrics mask cIoU.}
\resizebox{0.9 \textwidth}{!}{
   \tablestyle{5pt}{1.1}
    \begin{tabular}{l|c|ccc|ccc|cc}
    \toprule
                           &    & \multicolumn{3}{c|}{RefCOCO} & \multicolumn{3}{c|}{RefCOCO+} & \multicolumn{2}{c}{RefCOCOg} \\
    Models & avg & val & testA & testB & val & testA & testB & val & test \\
    \hline
MiniGPT-v2 & 62.42 & 72.63 & 65.30 & 65.07 & 66.22 & 60.68 & 58.37 & 54.94 & 56.18 \\

PAE-LaMM & 72.61 & 83.93 & 81.63 & 74.37 & 72.18 & 72.14 & 65.00 & 68.05 & 63.55 \\
\bottomrule
    \end{tabular}}
\end{table*}

\begin{figure}[!t]
    \centering
    \includegraphics[scale=0.50]{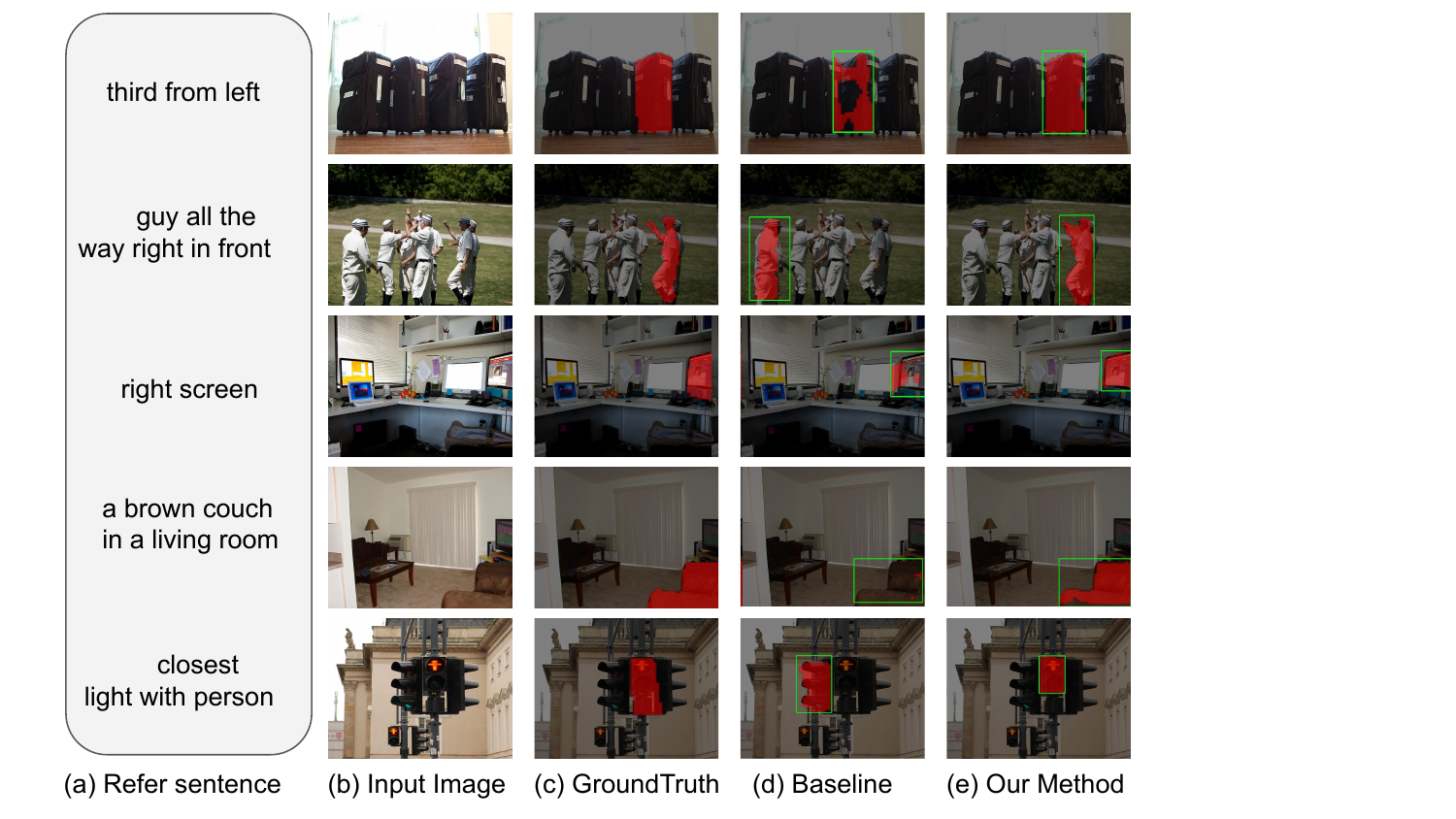}
   \caption{\textbf{Qualitative results of Referring Image Segmentation.} We first use the referring localization ability of the fine-tuned model to generate a bounding box (bbox) for the referring object, and then predict the segmentation mask inside the bbox.}
    \label{fig:seg}
\end{figure}

\noindent\textbf{Referring Image Segmentation}
For referring image segmentation, we report our results on a subset of RefCOCO~\cite{kazemzadeh2014referitgame}, RefCOCO+~\cite{yu2016modeling}, and RefCOCOg\cite{mao2016generation}. There are eight datasets in total: three for RefCOCO~\cite{kazemzadeh2014referitgame}, three for RefCOCO+~\cite{yu2016modeling}, and two for RefCOCOg~\cite{mao2016generation}. From each dataset, we select 100 data points as our subset. Following previous work~\cite{lai2023lisa} on referring image segmentation, we use cIoU as our evaluation metric, which is defined as the cumulative intersection over the cumulative union. ~\cref{tab:only2seg} shows our method outperforming the baseline model by a large margin. For example, our method improves baseline performance by $7.3$, $16.5$, and $9.5$ on val, testA, testB of RefCOCO, respectively. Because our fine-tuning uses both referring localization and segmentation data, we enable the model to first predict the bbox according to the referent sentence and then output the segmentation mask inside the bbox during inference.~\cref{fig:seg} shows the qualitative results of the baseline and our method; our method achieves better localization (the second row of ~\cref{fig:seg}) and pixel-level segmentation results (the bottom row of~\cref{fig:seg}). The significant improvement in referring image segmentation demonstrates the advantages brought by pixel reconstruction pretraining.

\begin{figure}[!t]
    \centering
    \includegraphics[scale=0.45]{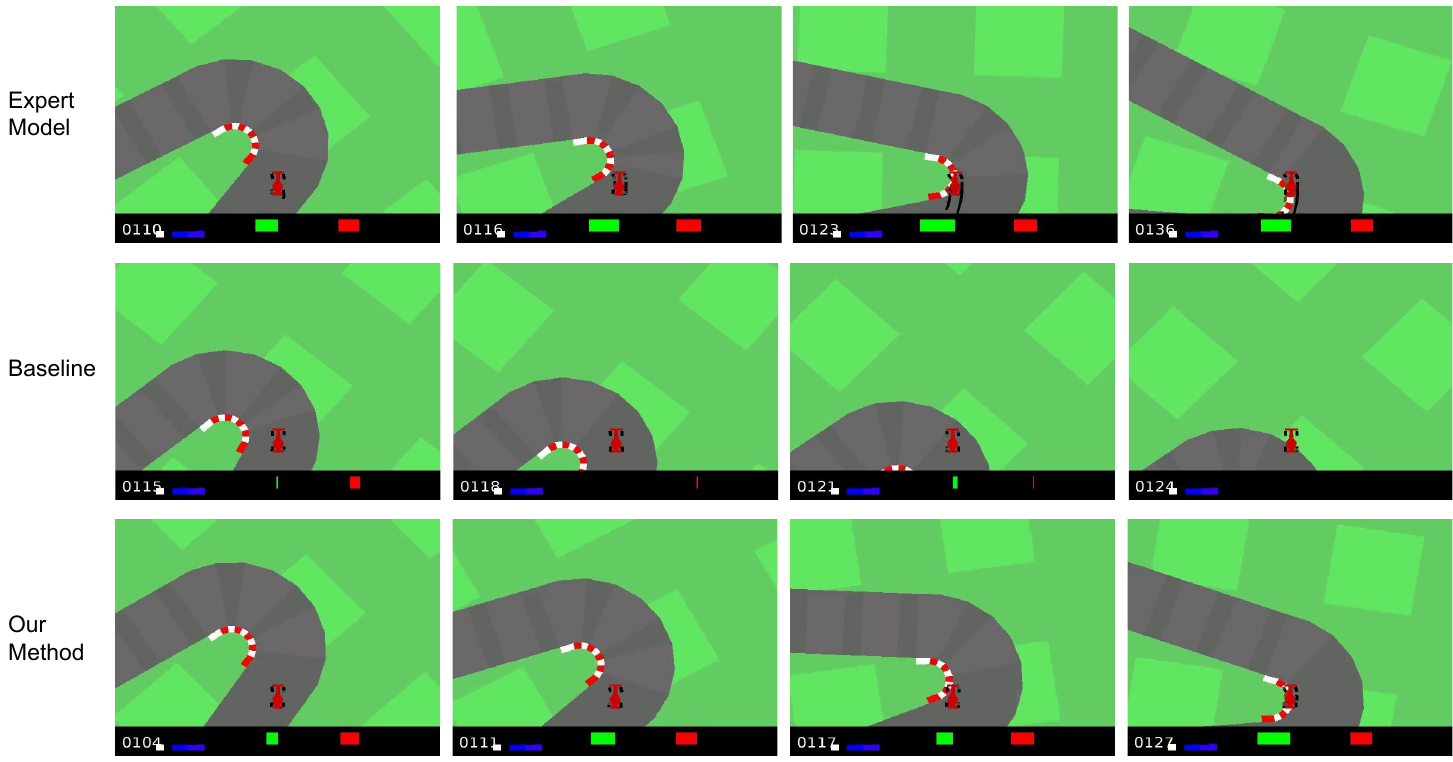}
   \caption{\textbf{Qualitative results of Carracing.} We show the game observation from different models, including the expert Reinforcement Learning (RL) Model, Baseline Model, and Our Method, all playing under the same game seed. These images depict how each model behaves when controlling the car and approaching the same corner.}
    \label{fig:carracing}
    
\end{figure}
\setlength{\tabcolsep}{4pt}
\begin{table}[t]
\scriptsize
\caption{\textbf{Results on Carracing and SpaceInvaders.} We report the average reward. The results for each game are collected in 15 rounds.}
\label{tab:game_playing}
\centering 
\resizebox{0.6\textwidth}{!}{
\begin{tabular}{@{}l|c|c @{}} %
\toprule 
\multirow{2}{*}{Method} & Carracing &SpaceInvaders\\
& mean reward & mean reward   \\
\hline
Expert Model  &853.91     &    476.33  \\ 
MiniGPT-v2 &465.36  &152.33  \\
\rowcolor{Gray}
PAE-LaMM &535.90  &232.67   \\
\bottomrule
\end{tabular}}

\end{table}

\noindent\textbf{Results on Video Game Playing.} We first report the game scores shown in ~\cref{tab:game_playing}. The score is obtained from the RL-Zoo3 library~\cite{rl-zoo3}. Here, we briefly introduce how the score is computed; details can be found in OpenAI Gym~\cite{openaiGym}. For CarRacing, the score is computed as -0.1 for every frame and +1000/N for every track tile visited~\cite{openaiGym}. For Space Invaders, players can gain points by destroying space invaders~\cite{openaiGym}. For both games, a higher score indicates better performance. 
As shown in~\cref{tab:game_playing}, for the CarRacing game, our method achieves a better mean reward over the baseline model with an 70.54 score gap. For the Space Invaders game, our method also outperforms the baseline with a score over 80.34. Additionally, we present the qualitative results of the CarRacing game in ~\cref{fig:carracing}, demonstrating how each model behaves when controlling the car and approaching the same corner. The expert model performs left steering and braking, and our method also predicts the action that makes the correct turn when approaching a sharp curve on the track. However, the baseline model fails to perform the correct action. Both the quantitative and qualitative results support our method in enhancing vision detail understanding.
\vspace{-3mm}
\subsection{Pixel Reconstruction Pre-training for VLM}
In~\cref{tab:vqa_results}, we show that our pre-trained model, PAE-LaMM, achieves results comparable to state-of-the-art models on multiple VQA tasks. Additionally, it has the ability to reconstruct pixels, which helps the VLM see image details. After fine-tuning for downstream referring sentence tasks, PAE-LaMM also achieves competitive results compared to other large VLM models on referring expression comprehension tasks without any extra model design, as shown in~\cref{tab:sota_det}. We then present an ablation study in~\cref{tab:ablation_for_seg} to analyze the impact of vision encoder adaptation and the pixel reconstruction task on vision detail awareness. We first train three models using three-stage pretraining under different settings. These models, along with a baseline model, are then fine-tuned on the referring segmentation task. The second row of~\cref{tab:ablation_for_seg} uses the same pretraining as PAE-LaMM while excluding the PVP task, and the third row freezes the vision encoder during the entire pretraining and includes the PVP task. The results show that both adapting ViT without the PVP task and fine-tuning the model on the PVP task without adapting ViT can improve the baseline result, with the improvement gap being almost the same. However, combining both adaptations leads to the best results, significantly outperforming each approach individually.
\setlength{\tabcolsep}{4pt}
\begin{table}[t]
\scriptsize
\caption{\textbf{Results on multiple VQA tasks.} We report top-1 accuracy for each task. Extra Ability indicates whether the model incorporates pixel-level reconstruction and visual localization capability. The best performance for each benchmark is indicated in \textbf{bold blue}, and the second best in blue.}
\label{tab:vqa_results}
\centering 
\resizebox{0.9\textwidth}{!}{
\begin{tabular}{@{}l|c c|ccccc @{}} %
\toprule
\multirow{2}{*}{Method} & \multicolumn{2}{c|}{Extra Ability} & \multicolumn{5}{c}{VQA tasks}\\
& Reconstruct & Grounding & GQA & VSR & IconVQA & VizWiz & HM \\  
\midrule
Flamingo-9B ~\cite{alayrac2022flamingo} & \xmark & \xmark & - & 31.8 & - & 28.8 & 57.0 \\

BLIP-2 (13B) ~\cite{blip2} & \xmark & \xmark & 41.0 & 50.9 & 40.6 & 19.6 & 53.7 \\

InstructBLIP (13B) ~\cite{dai2023instructblip} & \xmark & \xmark & 49.5 & 52.1 & 44.8 & 33.4 & \textcolor{blue}{57.5} \\ 
MiniGPT-4 (13B) ~\cite{minigpt4} & \xmark & \xmark & 30.8 & 41.6 & 37.6 & - & - \\
LLaVA (13B) ~\cite{llava} & \xmark & \xmark & 41.3 & 51.2 & 43.0 & - & - \\
Shikra (13B) ~\cite{shikra} & \xmark & \cmark & - & - & - & - & - \\
Qwen-VL (7B) ~\cite{bai2024qwenvl} & \xmark & \cmark & \textcolor{blue}{59.3} & - & - & 35.2 & - \\
MiniGPT-v2 (7B) ~\cite{minigptv2} & \xmark & \cmark & \textcolor{blue}{\textbf{60.1}} & \textcolor{blue}{\textbf{62.9}} & \textcolor{blue}{\textbf{51.5}} & 53.6 & \textcolor{blue}{\textbf{58.8}} \\
\rowcolor{Gray}
PAE-LaMM (7B) & \cmark & \cmark & 57.7 & \textcolor{blue}{59.2} & \textcolor{blue}{49.7} & \textcolor{blue}{\textbf{56.4}} & 57.2 \\
\bottomrule
\end{tabular}
}
\end{table}

\begin{table*}[t!]
\centering
\scriptsize
\caption{
        \label{tab:sota_det}
        \textbf{State-of-the-art comparison of LLM-based Methods on referring expression comprehension tasks.}
We report the official evaluation metrics: precision at IoU threshold $0.5$ for referring (box) localization. Numbers for other methods are taken from the original publications. The best performance for each benchmark is indicated in \textbf{bold blue}, and the second best in blue.
    }
\resizebox{1 \textwidth}{!}{
   \tablestyle{10pt}{1.0}
    \begin{tabular}{l|c|ccc|ccc|cc|c}
    \toprule
                               &  Extra    & \multicolumn{3}{c|}{RefCOCO} & \multicolumn{3}{c|}{RefCOCO+} & \multicolumn{2}{c}{RefCOCOg}  & \multicolumn{1}{c}{Average} \\
                               
    Models    &Decoder            & val     & testA   & testB   & val     & testA    & testB   & val           & test                                 \\
    \hline
VisionLLM\cite{wang2024visionllm}  & \xmark & 86.7          &      -         & -             & -             & -             & - & -             & -             & - \\      
Shikra-7B ~\cite{shikra} &\xmark  &   87.0          & 90.6          & 80.2          & 81.6          & \textcolor{blue}{87.4} & 71.1          & 82.3          & 82.2 & 82.8          \\
Ferret-7B ~\cite{you2023ferret}   &\xmark     & 87.5          & 91.4          & 82.5          & 80.8          & \textcolor{blue}{87.4} & 73.1          & 83.9          & 84.8   & 83.93       \\
Qwen-VL-7B ~\cite{bai2024qwenvl} & \xmark &\textcolor{blue}{89.36} &\textcolor{blue}{\textbf{92.26}} &85.34 &83.12 &\textcolor{blue}{\textbf{88.25}}& 77.21 &\textcolor{blue}{\textbf{85.58}} &85.48 & 85.83\\
MiniGPT-v2 ~\cite{minigptv2} & \xmark & 88.06 & 91.29 & 84.30 & 79.58 & 85.52 & 73.32 & 84.19 & 84.31 & 83.82 \\
PixelLLM ~\cite{xu2023pixel} &   \cmark & \textcolor{blue}{\textbf{89.8}} & \textcolor{blue}{92.2} & \textcolor{blue}{\textbf{86.4}} & \textcolor{blue}{83.2} & 87.0          & \textcolor{blue}{\textbf{78.9}} & 84.6 & \textcolor{blue}{86.0} & \textcolor{blue}{\textbf{86.01}} \\
\rowcolor{Gray}
PAE-LaMM& \xmark        & 88.55 & 91.1        & \textcolor{blue}{86.2} & \textcolor{blue}{\textbf{83.3}} & 87.30 & \textcolor{blue}{78.7} & \textcolor{blue}{85.2} & \textcolor{blue}{\textbf{86.6}} & \textcolor{blue}{85.87}\\
\bottomrule
    \end{tabular}}

\end{table*}

\begin{table*}[t!]
\scriptsize
\centering

\caption{
    \label{tab:ablation_for_seg}
    \textbf{Ablation on referring segmentation.} For each dataset, we test 100 referring sentences.
    We report the official evaluation metrics mask cIoU.}

\resizebox{0.85\textwidth}{!}{
   \tablestyle{5pt}{1.1}
    \begin{tabular}{l|c|ccc|ccc|cc}
    \toprule
                           &    & \multicolumn{3}{c|}{RefCOCO} & \multicolumn{3}{c|}{RefCOCO+} & \multicolumn{2}{c}{RefCOCOg} \\
    Models & avg & val & testA & testB & val & testA & testB & val & test \\
    \hline
PAE-LaMM & 72.61 & 83.93 & 81.63 & 74.37 & 72.18 & 72.14 & 65.00 & 68.05 & 63.55 \\
(-)  PVP task & 67.31 & 77.67 & 73.41 & 69.92 & 67.87 & 62.42 & 60.25 & 63.45 & 63.45 \\
(-) adapt ViT & 67.28 & 80.28 & 70.87 & 62.51 & 72.80 & 64.91 & 62.51 & 62.77 & 61.62 \\
Baseline & 62.42 & 72.63 & 65.30 & 65.07 & 66.22 & 60.68 & 58.37 & 54.94 & 56.18 \\
\bottomrule
    \end{tabular}}
    \vspace{-3mm}
\end{table*}

\setlength{\tabcolsep}{4pt}
\begin{table}[t]
\scriptsize
\caption{\textbf{Results of Multi-stage pretraining .} We report the Reconstruction Error, average referring expression comprehension (REC)
on refcoco, refcoco+ and refcocog. Also, we report the VQA task performance. The best performance for each benchmark is indicated in \textbf{bold}.}
\label{maintable:multi-stage-trains}
\centering 
\resizebox{0.9\textwidth}{!}{
\begin{tabular}{@{}l|c c c|ccccc @{}} %
\toprule
& adapt ViT & RE $\downarrow$ & Average REC $\uparrow$& GQA$\uparrow$ & VSR $\uparrow$ & IconVQA $\uparrow$& VizWiz $\uparrow$& HM $\uparrow$ \\ 

\midrule

Stage 1 & \xmark & 20.27 &76.8   & 55.5 & 56.3 & 49.5 & 40.1 & 55.2 \\

Stage 2 & \cmark& 6.65 &82.3 & 56.2 & 56.7 & 49.6 & 53.22 & \textbf{57.6} \\
Stage 3 & \xmark &\textbf{6.59} &\textbf{83.2} & \textbf{57.7} & \textbf{59.2} & \textbf{49.7} & \textbf{56.4} & 57.2 \\
\bottomrule
\end{tabular}
}

\end{table}
\setlength{\tabcolsep}{1.4pt}

In the end, ~\cref{maintable:multi-stage-trains} shows each stage performance of our three stage pretraining. Stage 3 achieves the best result in pixel reconstruction, REC, and most VQA tasks. The reconstruction error is significantly reduced after stage 2 (ViT adapting stage), which supports the notion that adapting ViT helps the VLM see more visual details. Additionally, the referring localization ability can be significantly improved by stage 2. Stage 3 mainly improves VQA tasks and REC performance while maintaining reconstruction ability. 
We provide more ablation studies in Supplementary material to demonstrate the training strategies of each stage.

\section{Conclusion}
In this paper, we mainly investigate the question, "How Well Can Vision Language Models See Image Details?". We propose a method to examine vision detail perception ability by querying VLMs to predict the pixels of input images given pixel locations. We find that VLMs struggle to reconstruct the original image using original CLIP vision features, and this issue can be significantly improved by adapting the vision encoder. We design a pixel reconstruction pretraining to enhance VLM vision detail perception ability. Then, we show the strong benefits brought by the ability to see image details in referring segmentation and playing video games. Additionally, we demonstrate that our pre-trained model, PAE-LaMM, does not lose general vision-language knowledge while possessing vision detail perception ability, which suggests our method may potentially be used for many vision-language applications, especially those requiring vision details.
\newpage

%
%
\bibliographystyle{splncs04}
\bibliography{bibtex}
\end{document}